%% file: main.tex
\documentclass[runningheads]{llncs}
\usepackage[T1]{fontenc}

\usepackage{xcolor}
\usepackage{graphicx}
\graphicspath{{supplement/}}
\usepackage{amsmath}
\usepackage{amssymb}
\usepackage{pifont}
\newcommand{\cmark}{\ding{51}}%
\newcommand{\xmark}{\ding{55}}%
\usepackage{tabu}
\usepackage{booktabs}
\usepackage{caption}
\usepackage{subcaption}

\usepackage{algpseudocode}
\usepackage{algorithm}

\usepackage{siunitx}
\usepackage{placeins}
\usepackage{ifthen}
\newboolean{my_hasappendix_check}
\usepackage{longtable}

\usepackage[misc,geometry]{ifsym}

\usepackage[pagebackref=true,breaklinks=true,colorlinks]{hyperref}
\usepackage{color}

\newcommand{\repeatthanks}{\textsuperscript{\thefootnote}}

\begin{document}
\setboolean{my_hasappendix_check}{true}

\title{Tracking by weakly-supervised learning and graph optimization for whole-embryo \textit{C. elegans} lineages}
\titlerunning{Tracking by learning and optimization}

\author{
  Peter Hirsch\inst{1,2}\orcidID{0000-0002-2353-5310}\Letter,\quad
  Caroline Malin-Mayor\inst{4}\orcidID{0000-0002-9627-6030},\quad
  Anthony Santella\inst{3},\quad
  Stephan Preibisch\inst{4}\orcidID{0000-0002-0276-494X},\quad
  Dagmar Kainmueller\thanks{equal contribution}\inst{1,2}\orcidID{0000-0002-9830-2415},\quad
  Jan Funke\repeatthanks\inst{4}\orcidID{0000-0003-4388-7783}%
}

\authorrunning{P. Hirsch et al.}
\institute{Max-Delbrueck-Center for Molecular Medicine in the Helmholtz Association, DE
  \email{\{peter.hirsch, dagmar.kainmueller\}@mdc-berlin.de}\and
  Humboldt-Universität zu Berlin, Faculty of Mathematics and Natural Sciences, DE \and
  \mbox{Sloan Kettering Cancer Center, Molecular Cytology Core, Developmental Biology, USA} \and
  HHMI Janelia Research Campus, USA
}

\maketitle              
%

\begin{abstract}
  Tracking all nuclei of an embryo in noisy and dense fluorescence microscopy data is a challenging task.
  We build upon a recent method for nuclei tracking that combines weakly-supervised learning from a small set of nuclei center point annotations with an integer linear program (ILP) for optimal cell lineage extraction. Our work specifically addresses the following challenging properties of \textit{C. elegans} embryo recordings: (1) Many cell divisions as compared to benchmark recordings of other organisms, and (2) the presence of polar bodies that are easily mistaken as cell nuclei. To cope with (1), we devise and incorporate a learnt cell division detector. To cope with (2), we employ a learnt polar body detector. We further propose automated ILP weights tuning via a structured SVM, alleviating the need for tedious manual set-up of a respective grid search.
  
  Our method outperforms the previous leader of the cell tracking challenge on the \textit{Fluo-N3DH-CE} embryo dataset.
  We report a further extensive quantitative evaluation on two more \textit{C.\ elegans} datasets.
  We will make these datasets public to serve as an extended benchmark for future method development.
 %
  Our results suggest considerable improvements yielded by our method, especially in terms of the
  correctness of division event detection and the number and length of fully correct track segments.
  Code: \url{https://github.com/funkelab/linajea}


\end{abstract}



\input{introduction}


\input{method}
\input{experiments}

\input{conclusion}


\subsubsection*{Acknowledgments}
We would like to thank Anthony Santella, Ismar Kovacevic and Zhirong Bao and Ryan Christensen, Mark W. Moyle and Hari Shroff for providing us with their data and annotations, for generously allowing us to make the data public  and for valuable information and feedback.
P.H. was funded by the MDC-NYU exchange program and HFSP grant RGP0021/2018-102. P.H. and D.K. were supported by the HHMI Janelia Visiting Scientist Program. A.S. was supported by grant 2019-198110 (5022) from the Chan Zuckerberg Initiative and the Silicon Valley Community Foundation.

\clearpage
{
  \bibliographystyle{splncs04}
  \bibliography{main}
}

\ifthenelse{\boolean{my_hasappendix_check}}{
\clearpage
\appendix
\input{supplement/supplement_content}
}{}

\end{document}

%% file: introduction.tex
\section{Introduction}

\begin{figure}[htbp]
  \centering
  \includegraphics[width=\textwidth]{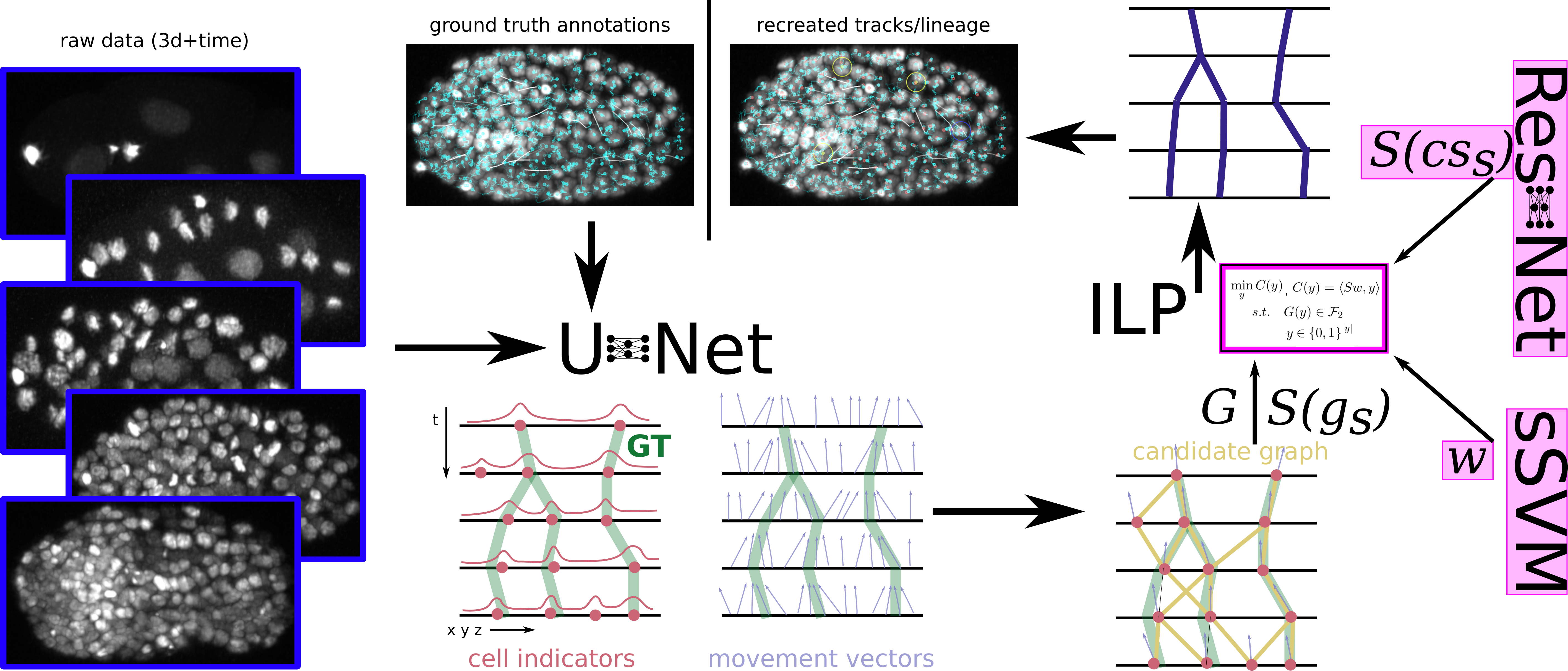}
  \caption{\label{fig:method} Method overview: We use a 4d U-Net to predict cell candidates and movement vectors. These are used to construct a candidate graph \(G\) with node and edge scores \(g_s\) as in~\cite{Malin-Mayor2021_autom_recon_of_whole_embry_cell_linea_by_learn_from_spars_annot}. We propose to integrate learnt cell state scores \(cs_s\). Graph \(G\), feature matrix \(S\),  weights \(w\) and a set of feasibility constraints form an ILP that yields the cell lineage. We propose to find \(w\) via a structured SVM (sSVM); proposed changes highlighted in magenta; figure adapted from~\cite{Malin-Mayor2021_autom_recon_of_whole_embry_cell_linea_by_learn_from_spars_annot}.
  }
\end{figure}

Advances in microscopy have made the recording of whole embryo development possible~\cite{keller2010fast,krzic2012multiview}.
%
However there is an inherent trade-off between frame rate, resolution and the prevention of phototoxicity~\cite{Weigert2018_CARE}.
While it is possible to capture high signal-to-noise images with high resolution, this can damage the organism, especially during early embryonic development.
Consequently, embryonic development is commonly captured at low frame rate and with low signal-to-noise ratio (SNR). Together with cell-cycle inherent signal fluctuation, low SNR renders automated cell detection challenging. 
Furthermore, low frame rate renders automated tracking challenging as overlap-based tracking approaches are not applicable. Last but not least, the similar shape and appearance of distinct cell nuclei also renders similarity-based tracking approaches ineffective.
%

A number of automated cell tracking approaches have been developed to tackle reduced SNR as well as frame rates on the order of minutes.
Such methods have enabled a range of studies on a variety of organisms, where it wouldn't have been feasible to do tracking manually
\cite{li2019_syste_probe_and_spati_regul_of_cell_posit_varia_durin_embry,murray08_autom_analy_embry_gene_expres,cao2020establishment,medeiros2021_multisc_light_sheet_organ_imagi_frame,Wolff2018,guignard20_conta_area_depen_cell_commu_and_the_morpho_invar_of_ascid_embry}.
%
The \emph{Cell Tracking Challenge} (CTC)~\cite{ulman17_objec_compar_cell_track_algor}, an extensive benchmark that contains 2d+time and 3d+time datasets of different organisms recorded with a variety of microscopes, allows for a quantitative comparison of automated cell tracking methods.

Current methods for cell tracking in \textit{C.\ elegans} follow the \emph{tracking-by-detection} paradigm, which first computes (candidate) cell detections in all frames, and in a second step links matching cell detections across frames.
To this end, Starrynite~\cite{bao06_autom_cell_lineag_tracin_caenor_elegan}, which is widely used by practitioners, uses classical computer vision to detect locations of maximum signal in each frame, nearest neighbor matching for link detection, and local post-processing to resolve ambiguities that occur in case of cell divisions. 
Linkage can also be achieved in a globally optimal manner by means of combinatorial optimization, as in the competitive method Baxter~\cite{magnusson15_global_linkin_cell_track_using_viter_algor} (former rank 2 on the CTC \textit{C.\ elegans} benchmark), which employs the Viterbi algorithm in the linkage step.  
Related methods can also directly yield an optimal feasible cell lineage tree, in the face of over- and underdetections~\cite{schiegg2013conservation}, as well as for an overcomplete set of candidate detections~\cite{jug2014optimal,Malin-Mayor2021_autom_recon_of_whole_embry_cell_linea_by_learn_from_spars_annot}.
To modernize the detection step, Cao et al.\ \cite{cao2020establishment} replaced the classical detection step of Starrynite by neural network-based cell segmentation.
%
Similarly, the recently proposed active learning framework Elephant~\cite{sugawara2021_track_cell_linea_in_3d_by_incre_deep_learn} (former rank 1 on the CTC \textit{C.\ elegans} benchmark) uses modern deep learning for segmentation and detection. However, both perform local linkage via nearest neighbor search (in case of Elephant optionally supported via a learned optical flow estimate).

We propose a method that unifies the individual advantages of Baxter~\cite{magnusson15_global_linkin_cell_track_using_viter_algor} and Elephant~\cite{sugawara2021_track_cell_linea_in_3d_by_incre_deep_learn}.
To this end we build upon Linajea~\cite{Malin-Mayor2021_autom_recon_of_whole_embry_cell_linea_by_learn_from_spars_annot}, a recent method that combines deep learning and combinatorial optimization for cell tracking.
We propose extensions of Linajea to capture properties specific to recordings of the model organism \emph{C.\ elegans}, namely relatively many cell divisions, and the presence of \emph{polar bodies}
which look similar to nuclei. Our extended method yields state of the art accuracy on benchmark \emph{C.\ elegans} data.
Besides accuracy, we also address efficiency.
We propose to use a structured SVM (sSVM) to facilitate the tuning of the weights of the ILP objective which alleviates the need for manual configuration of a grid search.
This is particularly useful in light of our extensions as they introduce additional weights and thereby would increase the dimensionality of the search.
In summary our contributions are:
\begin{itemize}
\item A learnt cell state and polar body detector, integrated into an existing approach that combines deep learning and an ILP for nuclei tracking.
\item Fully automated tuning of the weights of the ILP objective via a sSVM.
\item The new state-of-the-art for \emph{C.\ elegans} in the Cell Tracking Challenge.
\item Two new datasets made publicly available as benchmark data, namely three con\-fo\-cal and three lightsheet recordings of \textit{C.\ elegans}, all fully annotated.
\end{itemize}


%% file: method.tex
\section{Method}

Our method extends the tracking-by-detection approach Linajea~\cite{Malin-Mayor2021_autom_recon_of_whole_embry_cell_linea_by_learn_from_spars_annot}.
We briefly review Linajea, and then describe our extensions in detail. For more details on Linajea, please refer to~\cite{Malin-Mayor2021_autom_recon_of_whole_embry_cell_linea_by_learn_from_spars_annot}.
For an overview of our extended method, see Fig.~\ref{fig:method}.

\paragraph{\textbf{Linajea. }}
\label{subsec:linajea}
Linajea~\cite{Malin-Mayor2021_autom_recon_of_whole_embry_cell_linea_by_learn_from_spars_annot} implements a 4d U-Net~\cite{ronneberger15_u_net,ccicek16_u_net,funke_conv4d} to predict the position and movement of each nucleus.
Position is encoded as a single-channel image of Gaussian-shaped blobs, one per nucleus, as in~\cite{hofener2018deep}.
The locations of the respective intensity maxima correspond to nuclei center points. 
Movement is encoded as 3d vectors per pixel within a nucleus. Each vector points to the spatial location (center point) of the same (or parent) nucleus in the previous time frame.
Note, the backwards direction of the movement vectors simplifies tracking as cells can only divide going forward but cannot merge.
The four output channels necessary for the above encoding are trained jointly via L2 loss. 

During inference, local maxima of the predicted position map serve as cell candidates. 
An integer linear program (ILP) is employed to select and link a feasible subset of these cell candidates. To this end, a candidate graph is established, where nodes represent cell candidates, and edges represent cell linkage candidates. Node- and edge costs are derived from the U-Net's position- and movement prediction channels, respectively.
%
%
Solving the ILP assigns a label "selected" or "not selected" to each candidate cell (node) and each candidate link (edge) such that a cost-based objective is minimized. Linear constraints on the node- and edge labels ensure that a valid tracking solution is extracted from the graph, i.e., a binary forest where each tree only branches forward in time.

Linajea's objective is a weighted sum of the costs of the selected nodes and edges. There are four tunable weights: A constant cost \(w_{\text{node-sel}}\) for selecting a node, a factor \(w_{\text{node-cost}}\) to scale the position prediction, a factor \(w_{\text{edge-cost}}\) to scale the distance between predicted movement vector target and linked cell position, and a constant cost \(w_{\text{track-cost}}\) for each track. The track cost is incorporated via an additional binary node label "track start", with appropriate constraints to ensure consistency with the selection labels.
Linajea performs grid-search within a manually defined range to find a set of suitable weights.


We propose two extensions to Linajea: (1) An additional network to classify cell state,
and respective additional costs and feasibility constraints in the ILP, and (2) the use of a structured SVM (sSVM) to automatically find the optimal weights for the ILP objective, as described in the following.

\paragraph{\textbf{Cell State Classifier. }}
\label{subsec:cell_state}
We propose to incorporate a classifier to determine the cell state of each cell candidate similarly to~\cite{santella14_semi_local_neigh_based_framew}.
We assign to each candidate one of four classes: parent cell (i.e., a cell that is about to undergo cell division), daughter cell (cell that just divided), continuation cell (cell track that continues without division) and polar body.
We train a 3d ResNet18~\cite{he15_deep_resid_learn_image_recog} with 3d convolutions for this task.
The parent/daughter/continuation classes are incorporated into the ILP as a separate set of node labels, with their weighted prediction scores as costs.
%
In addition to Linajea's feasibility constraints on node and edge selection~\cite{Malin-Mayor2021_autom_recon_of_whole_embry_cell_linea_by_learn_from_spars_annot}, we impose novel constraints that ensure that (1) selection- and cell state labels are consistent, and (2) a parent at time \(t\) can only be linked to a daughter at time \(t\)+1 and vice versa.

Formally, for each edge \(e=(u,v)\) between nodes \(u\) and \(v\) in the graph (directed forward in time), let \(y_{\text{edge},e}, \ y_{\text{node},u} \in \{0,1\}\) denote binary variables that represent edge- and node selection as in Linajea. We introduce novel binary variables \(y_{\text{daughter},u}\), \(y_{\text{parent},u}\), \(y_{\text{continue},u}\) \(\in \{0,1\}\) that represent daughter-, parent- and continuation cell state labels per node.
To ensure that a selected node is assigned exactly one cell state, for each node, we introduce the linear equality constraint \(y_{\text{parent},u} + y_{\text{daughter},u} + y_{\text{continue},u} - y_{\text{node},u} = 0\) to be included in the ILP. To ensure that selected edges constitute feasible parent-daughter links, for each edge, we introduce novel  inequality constraints \( y_{\text{parent},u} + y_{\text{edge},e} - y_{\text{daughter},v} \leq 1\) and \( y_{\text{daughter},v} + y_{\text{edge},e} - y_{\text{parent},u} \leq 1\) to be included in the ILP. Thus, e.g., if node \(u\) is labelled parent (\(y_{\text{parent},u} =1\)), and edge \(e\) is selected (\(y_{\text{edge},e} =1\)), node \(v\) has to be labelled daughter (\(y_{\text{daughter},v} =1\)). 

We add the cell state predictions to Linajea's objective. This entails new weights \(w_{\text{parent}}, w_{\text{daughter}}, w_{\text{continue}}\) that serve to scale the prediction scores of respectively selected and labelled nodes. A further weight \(w_{\text{division}}\) serves as constant division cost, contributed by each selected parent.


By default, we do not perform any postprocessing on the tracks (such as removal of short tracks).
We propose one exception regarding the polar bodies:
Depending on the specific study they might not be of interest, and even if, they are often not contained in the ground truth tracks as they are not considered ``proper'' cells.
That is why we add them as an additional class to our cell state classifier.
The score for the polar body class can be used to optionally detect and remove them from the tracks.  Suppl.~Fig.~\ref{fig:pol_apo_bodies} shows an exemplary polar body.

\paragraph{\textbf{Structured SVM-based weights search.}}
\label{subsec:ssvm}

Manually configured grid search for optimal weights as in~\cite{Malin-Mayor2021_autom_recon_of_whole_embry_cell_linea_by_learn_from_spars_annot} can be costly and generally a dataset specific search range has to be found for each new dataset.
To alleviate the need for this manual step, we propose the use of a structured SVM (sSVM) for automatic weight selection~\cite{joachims09_predi_struc_objec_with_suppo_vecto_machin,teo10_bundl_metho_for_regul_risk_minim}.
Following~\cite{Malin-Mayor2021_autom_recon_of_whole_embry_cell_linea_by_learn_from_spars_annot}, our extended objective can be phrased as
\begin{equation}
  \min_{\mathbf{y}} \langle S\mathbf{w},\mathbf{y} \rangle \quad s.t. \quad G(\mathbf{y}) \in \mathcal{F}_2,
  \label{eq:ilp_cost}
\end{equation}
 where \(\mathbf{y}\) is
%
 a vector of all binary indicator variables, including our new ones 
%
\[\mathbf{y} = \big[\mathbf{y}_{\text{node}}^T,\mathbf{y}_{\text{track}}^T,\mathbf{y}_{\text{parent}}^T,\mathbf{y}_{\text{daughter}}^T,\mathbf{y}_{\text{continue}}^T, \mathbf{y}_{\text{edge}}^T\big]^T \in \{0, 1\}^{5|V|+|E|}. \]
\(G(\mathbf{y})\) denotes the graph formed by the selected nodes and edges in \(\mathbf{y}\). \(\mathcal{F}_2\) denotes the set of all feasible binary forests.
\(S^{\text{dim}(y)\times \text{dim}(w)}\) is a sparse feature matrix that contains all node- and edge features. It has the following columns:
(1) 1 for \textit{node} indicators (i.e., in the first \(|V|\) rows), and 0 otherwise.
(2) Candidate cell prediction for \textit{node} indicators, and 0 otherwise.
(3) 1 for \textit{track start} indicators, 0 otherwise.
(4) 1 for \textit{parent class} indicators, 0 otherwise.
(5-7) Parent/daughter/continue class predictions for \textit{parent/daughter/continue} indicators, 0 otherwise.
(8) Edge cost 
for \textit{edge} indicators, and 0 otherwise.
%
\(\mathbf{w}\) is the vector of weights
\[
  \mathbf{w} = \big[w_{\text{node-sel}}, w_{\text{node-score}}, w_{\text{track}}, w_{\text{div}}, w_{\text{parent}}, w_{\text{daughter}}, w_{\text{continue}}, w_{\text{edge}} \big]^T.
\]


Solving the ILP~\eqref{eq:ilp_cost} yields the best feasible \(\mathbf{y}\) given some \(\mathbf{w}\).
However, appropriate values for \(\mathbf{w}\) are unknown a priori.
With the help of the ground truth annotations, what we can determine though is a ``best effort'' indicator vector \(\mathbf{y}'\).
This equates to the best possible feasible solution given the set of predicted cell candidates and movement vectors.
We thus want to find the weights \(\mathbf{w}\) such that solving \eqref{eq:ilp_cost} yields \(\mathbf{y}=\mathbf{y}'\), or as close as possible to it.

To find such weights, given \(\mathbf{y}'\), we derive a modified objective from Eq.~\ref{eq:ilp_cost} which we then minimize w.r.t.\ the weights. 
We thus follow the sSVM approach with a loss-augmented objective~\cite{joachims09_predi_struc_objec_with_suppo_vecto_machin,teo10_bundl_metho_for_regul_risk_minim}. Formally, we seek a  \(\mathbf{w}\) that minimizes
\begin{equation} L(\mathbf{w}) = \langle S\mathbf{w},\mathbf{y}'\rangle - \min_{\mathbf{y}:\ G(\mathbf{y}) \in \mathcal{F}_2} \big( \langle S\mathbf{w},\mathbf{y}\rangle - \Delta(\mathbf{y}',\mathbf{y}) \big) + \lambda  | \mathbf{w} | ^2 \ , 
\end{equation}
with Hamming cost function \(\Delta\) to measure the deviation of the optimal \(\mathbf{y}\) for a given \(\mathbf{w}\) and the best effort \(\mathbf{y}'\), and a hyperparameter \(\lambda\geq 0\) for weighing L2 regularization on the weights (we use \(\lambda=0.001\)).

To give a brief intuition why optimizing this loss yields the desired parameters (for which we neglect the L2 regularizer for the moment): It is easy to see that \(L(\mathbf{w}) \geq 0\) because \(\Delta \geq 0\) 
and
\(\min_{\mathbf{y}:\ G(\mathbf{y}) \in \mathcal{F}_2}  \langle S\mathbf{w},\mathbf{y}\rangle  \leq \langle S\mathbf{w},\mathbf{y}'\rangle\). 
Furthermore, if \(\mathbf{w}\) yields \(\mathbf{y}'\) as minimum of the ILP, \(\arg\!\min_{\mathbf{y}:\ G(\mathbf{y}) \in \mathcal{F}_2}  \langle S\mathbf{w},\mathbf{y}\rangle = \mathbf{y}'\), then \(L(\mathbf{w}) = 0\), i.e., the loss is minimized. Last but not least, if a \(\mathbf{w}\) with zero loss does not exist, the loss seeks a \(\mathbf{w}\) that yields an ILP-minimizing \(\mathbf{y}\) that is at least ``close'' to \(\mathbf{y}'\) both in terms of the Hamming loss and in terms of its objective value \(\langle S\mathbf{w},\mathbf{y}\rangle\).
For details on the sSVM optimization procedure, please refer to~\cite{joachims09_predi_struc_objec_with_suppo_vecto_machin,teo10_bundl_metho_for_regul_risk_minim}.

%% file: experiments.tex
\section{Results}
\label{sec:exp}

To measure the performance of our method we evaluate it on three different datasets of developing \textit{C. elegans} embryos, the \textbf{Fluo-N3DH-CE} dataset of the Cell Tracking Challenge benchmark (CTC)~\cite{ulman17_objec_compar_cell_track_algor}, three confocal recordings (\textbf{mskcc-confocal}) and three lightsheet recordings (\textbf{nih-ls}).
See Suppl.\ Table~\ref{suppl_tab:impl} for information on implementation and computational details.

\paragraph{The \textbf{Fluo-N3DH-CE} dataset}
\label{subsec:ctc_data}
~\cite{murray08_autom_analy_embry_gene_expres} consists of four 3d+time anisotropic confocal recordings until the 350 cell stage; 2 public ones for training and 2 private ones for the official evaluation.
All tracks are annotated.
The polar body filter is not used for this dataset.
Our method (named JAN-US) achieves a detection score (DET) of 0.981, and a tracking score (TRA) of 0.979, thereby outperforming the previous state of the art from  Elephant~\cite{sugawara2021_track_cell_linea_in_3d_by_incre_deep_learn} (DET 0.979, TRA 0.975) at the time of submission. These results, which are listed on the challenge website, were generated using our cell state classifier (\textit{linajea+csc}).
%
See Suppl. Table~\ref{suppl_tab:metrics} for a short description of the metrics.
For details on the dataset, the challenge format and the metrics, please refer to~\cite{ulman17_objec_compar_cell_track_algor,matula15_cell_track_accur_measur_based}.

\textbf{Discussion. } As the labels for CTC test data are not public, a qualitative assessment of the improvement over the previous state of the art is not possible. However, the challenge scores are defined to be interpretable in terms of reduction of manual labor necessary for fixing an automated tracking solution. In this regard, our improvement in TRA over Elephant should mean that our method entails a 16\% reduction in manual curation effort as compared to Elephant (\(\frac{(1-0.975)-(1-0.979)}{(1-0.975)}=0.16\)).

Our improvement in DET and TRA scores on the challenge test data can further be put into perspective by comparison with the improvements in DET and TRA that we obtain and analyze on our other datasets (as described in the following, summarized in Table~\ref{tab:nuclei_results}).
We take this comparison as further indication that in terms of DET and TRA, an improvement in the 3rd decimal place as achieved on the CTC data can mean a considerable difference in performance. 
%
\begin{table*}[tbp]
\begin{center}
  \captionof{table}{Quantitative results on \textbf{mskcc-confocal} and \textbf{nih-ls} data. For description of error metrics see Suppl.\ Table~\ref{suppl_tab:metrics}; absolute number of errors normalized per 1000 GT edges; best value bold, value with insignificant difference to best value underlined (significance assessed with Wilcoxon's signed-rank test and \(p<0.01\)).}
    \begin{tabu} to 1.0\linewidth{ X[2.8l] | X[1.0c] X[1.0c] X[1.0c] X[1.2c] X[1.2c] | X[1.2c] | X[1.0c] | X[1.6c] | X[1.6c] }
      \toprule
       & FP & FN & IS & FPdiv & FNdiv & div & sum &  \multicolumn{1}{c|}{DET} & TRA\\
      \toprule
      \multicolumn{10}{c}{\textbf{mskcc-confocal} 270 frames}\\
      \midrule
      Starrynite
      & 7.9&	13&	0.62&	0.58&	1.2&	1.8&	24 & 0.97875 & 0.97495\\
      linajea
      & \textbf{3.6}&	\textbf{5.5}&	\underline{0.062}&	0.89&	\textbf{0.26}&	1.2&	10.3 & 0.99514 & 0.99418\\
      lin.+csc+sSVM
      & \underline{3.7}&	\underline{5.6}&	\textbf{0.046}&	\textbf{0.053}&	0.40&	\textbf{0.46}&	\textbf{9.6} & \textbf{0.99570} & \textbf{0.99480}\\
      \midrule
      \multicolumn{10}{c}{\textbf{nih-ls} 270 frames}\\
      \midrule
      Starrynite
      & 22 & 18 & 2.4 & 0.66 & 1.6 & 2.2 & 45 & 0.81850 & 0.81114  \\
      linajea
      &\textbf{12} &	\underline{6.5}&	\textbf{0.46}&	1.5 &	\textbf{0.40}&	1.86&	\underline{21} & 0.99367 & 0.99279\\
      lin.+csc+sSVM
      &\underline{13} &	\textbf{5.3}& 	\underline{0.59}&	\textbf{0.20}&	\underline{0.49}&	\textbf{0.69}&	\textbf{20} & \textbf{0.99511} & \textbf{0.99433}\\
    \end{tabu}
    \label{tab:nuclei_results}

  \end{center}
\end{table*}

\paragraph{The \textbf{mskcc-confocal} dataset}
\label{subsec:conf_data}
consists of three fully annotated 3d+time an\-iso\-tro\-pic confocal recordings (data: \url{https://doi.org/10.5281/zenodo.6460303}).
The ground truth has been created using Starrynite~\cite{bao06_autom_cell_lineag_tracin_caenor_elegan,santella14_semi_local_neigh_based_framew}, followed by manual curation (supported and verified by using the fixed \textit{C. elegans} lineage).
The annotations include the polar bodies (marked separately).
We report the uncurated Starrynite results as a baseline.
As we perform weakly-supervised training on point annotations yet Elephant requires segmentation masks we cannot compare to it on this data.
We train and evaluate all models on the first 270 frames (approximately 570 cells in the last frame and 52k in total per sample). 

%

Per experiment we use one recording each as training, validation and test set.
We do this for all six possible combinations. For each combination, we perform three experimental runs, starting from different random weight initializations, leading to a total of 18 experimental runs (all numbers averaged).
%
Divisions that are off by one frame compared to the annotations are not counted as errors as the limited frame rate leads to inherent inaccuracies in the data and annotations.

Both Linajea and our extended method considerably outperform Starrynite (see Table~\ref{tab:nuclei_results}, and Suppl.\ Fig.\ \ref{fig:error_plot} for the respective box plots).
We conducted an ablation study on the \textbf{mskcc-confocal} dataset to measure the effect of the individual extensions we propose (see Table~\ref{suppl_tab:ablation}):
%
We report results without the ILP, and without the cell state classifier in the ILP (this matches~\cite{Malin-Mayor2021_autom_recon_of_whole_embry_cell_linea_by_learn_from_spars_annot}).
Both strongly suffer from false positive (FP)-type errors.
We compare results with and without sSVM for weights tuning, and find that sSVM-determined weights yield competitive results.
The sSVM finds similar weights for all experimental runs (see Suppl. Fig.\ \ref{fig:ssvm_param_dist}).
Finally we employ the polar body filter (we remove them from the ground truth, too).
This reduces FP errors.
Due to the strong tree structure of the tracks the notion of ``how many tracks are correct'' is not well defined.
To still try to quantify the intention behind it we evaluate the fraction of error-free tracklets of varying lengths~\cite{Malin-Mayor2021_autom_recon_of_whole_embry_cell_linea_by_learn_from_spars_annot}, see Suppl.\ Fig.\ \ref{fig:corr_segments}.

\noindent\textbf{Discussion. } We did not expect to see large differences between the results for the sSVM-de\-ter\-mined weights and for the manually configured grid search as we have gathered experience in choosing appropriate parameters for the weights grid search for this data.
Thus the explicit search is often faster as it can be parallelized trivially.
However, for other data, where this information is not at hand, the targeted sSVM is very convenient and is computationally more efficient.
Interestingly, depending on the weights, the system appears to be able to exchange FP and FN errors.
The sSVM-determined weights seem to prioritize FP errors.
By adapting the cost function \(\Delta\) one can modulate this depending on respective application-specific needs (see Suppl.\ Table~\ref{suppl_tab:delta_res}).
  \begin{table*}[tbp]
  \centering
  \caption{Ablation study on 
  the \textbf{mskcc-confocal} data. We ablate solving an ILP altogether (ILP), incorporating the cell state classifier (csc), employing an sSVM for weights search (ssvm), and incorporating the polar body filter (pbf).}
    \begin{tabu} to 1.0\linewidth{ X[0.5c] X[0.5c] X[0.5c] X[0.7c] | X[0.5c] X[0.5c] X[0.9c] X[1.0c] X[1.0c] | X[1.0c] | X[0.7c] | X[1.2c] | X[1.2c]}
      \toprule
      ILP & csc & ssvm & pbf & FP & FN & IS & FPdiv & FNdiv & div & \multicolumn{1}{c|}{sum}  & DET & TRA\\
      \toprule
      \multicolumn{12}{c}{\textbf{mskcc-confocal} 270 frames}\\
      \midrule
      \xmark & \xmark & \xmark & \xmark &
      5.0&	\textbf{4.6}&	\underline{0.048}&	1.6&	\textbf{0.25}&	1.9&	11.6 & \underline{0.99567}& 0.99464 \\
      \cmark & \xmark & \xmark & \xmark &
      \underline{3.6}&	5.5&	0.062&	0.89&	\underline{0.26}&	1.2&	10.3 & 0.99514 & 0.99418\\
      \cmark & \cmark & \xmark & \xmark &
      \underline{3.4}&	5.7&	\textbf{0.028}&	0.11&	\underline{0.27}&	\textbf{0.38}&	\underline{9.5} & 0.99526 & 0.99437\\
      \cmark & \cmark & \cmark & \xmark &
      \underline{3.7}&	5.6&	\underline{0.046}&	\underline{0.053}&	0.40&	0.46&	\underline{9.6} & \textbf{0.99570} & \textbf{0.99480}\\
      \cmark & \cmark & \cmark & \cmark &
      \textbf{2.5}&	5.5&	\underline{0.047}&\textbf{0.048}&	0.39& 0.44 &\textbf{8.5} & \underline{0.99533} & \underline{0.99441}
\\
      \bottomrule
    \end{tabu}
    \label{suppl_tab:ablation}
  \end{table*}

\paragraph{The \textbf{nih-ls} dataset}
\label{subsec:ls_data}
contains three fully annotated 3d+time isotropic lightsheet recordings~\cite{Moyle2021} (data: \url{https://doi.org/10.5281/zenodo.6460375}).
Our experimental setup is similar to \textbf{mskcc-confocal}.

\noindent\textbf{Discussion. }It is interesting to compare our results on \textbf{nih-ls} and \textbf{mskcc-confocal}:
Due to the isotropic resolution of \textbf{nih-ls} we expected the results to be superior, yet the error metrics we observe do not support this intuition.
A closer look at qualitative results reveals some clues that may explain part of it:
Apoptotic cells are more distinct and visible earlier in \textbf{nih-ls} (see e.g. Suppl.\ Fig.\ \ref{fig:pol_apo_bodies}) and thus have not been annotated in the ground truth.
Yet in the current state our model does not handle this transition explicitly and thus continues to track them temporarily, leading to a larger number of false positives, as indicated by the quantitative results.
As we already have a cell state classifier as part of our model, it will be straightforward to add apoptotic cells as a remedy.




%% file: conclusion.tex
\section{Conclusion}

We presented extensions to the tracking method Linajea~\cite{Malin-Mayor2021_autom_recon_of_whole_embry_cell_linea_by_learn_from_spars_annot} to improve tracking of all cells during embryonic development.
In addition to combining deep learning to learn position and movement vectors of each cell and integer linear programming to extract tracks over time and ensure long term consistency, we integrate cell state information into the ILP, together with a method to automatically determine the weights of the ILP objective, alleviating the need for potentially suboptimal manually configured grid-search.

At the time of submission our method headed the leaderboard of the CTC for the DET and TRA scores for the \textbf{Fluo-N3DH-CE} dataset.
On two other datasets of both confocal and lightsheet recordings of \textit{C. elegans} our method outperforms the tool Starrynite, which is often used by practitioners for studies of \textit{C. elegans}, by a wide margin. Furthermore, an ablation study reveals that each of our proposed methodological advances improves upon baseline Linajea.

The low error rate achieved by our method
will
further
push down the required time for manual curation
This will facilitate studies that require a large number of samples.
More effort is still necessary in the later stages of development.
In future work we will extend the tracking all the way to the end of the embryonic development.
This poses additional challenges as the whole embryo starts to twitch, causing abrupt movements.
A second avenue of future work is to combine the two stages of the method.
Recent work~\cite{pogan2020_diffe_of_black_combi_solve} has proposed a method to incorporate black box solvers into a gradient-based end-to-end neural network learning process.
This shows great promise to increase the performance of our method even further.


%% file: supplement/supplement_content.tex
\clearpage
\title{Tracking by weakly-supervised learning and graph optimization for whole-embryo \textit{C. elegans} lineages:
Supplementary Material}
\titlerunning{Tracking by learning and optimization}

\author{
  Peter Hirsch\inst{1,2}\orcidID{0000-0002-2353-5310}\Letter,\quad
  Caroline Malin-Mayor\inst{4}\orcidID{0000-0002-9627-6030},\quad
  Anthony Santella\inst{3},\quad
  Stephan Preibisch\inst{4}\orcidID{0000-0002-0276-494X},\quad
  Dagmar Kainmueller\thanks{equal contribution}\inst{1,2}\orcidID{0000-0002-9830-2415},\quad
  Jan Funke\repeatthanks\inst{4}\orcidID{0000-0003-4388-7783}%
}

\authorrunning{P. Hirsch et al.}
\institute{Max-Delbrueck-Center for Molecular Medicine in the Helmholtz Association, DE
  \email{\{peter.hirsch, dagmar.kainmueller\}@mdc-berlin.de}\and
  Humboldt-Universität zu Berlin, Faculty of Mathematics and Natural Sciences, DE \and
  \mbox{Sloan Kettering Cancer Center, Molecular Cytology Core, Developmental Biology, USA} \and
  HHMI Janelia Research Campus, USA
}
\maketitle

\setcounter{figure}{1}
\setcounter{table}{2}
\section{Appendix}
\label{sec:appendix}
\FloatBarrier
\begin{center}
  \captionof{table}{Quantitative results on \textbf{mskcc-confocal} data with modified \(\Delta\) cost function. The cost for each FP/FN is multiplied by 10/100 respectively.}
  \begin{tabu} to 1.0\linewidth{ X[2.8l] | X[1.0c] X[1.0c] X[1.0c] X[1.2c] X[1.2c] | X[1.2c] | X[1.0c] }
    \toprule
    & FP & FN & IS & FPdiv & FNdiv & div & sum\\
    \toprule
    \multicolumn{8}{c}{\textbf{mskcc-confocal} 270 frames}\\
    \midrule
    lin.+csc+sSVM
    & 3.7&	5.6&	0.046&	0.053&	0.40&	0.46&	9.6\\
    \midrule
    +FN*10
    & 3.8&	5.1&	0.11&	0.33&	0.30&	0.62&	9.6\\
    +FN*100
    & 4.4&	4.9&	0.14&	0.95&	0.24&	1.1&	10.6\\
    \midrule
    +FP*10
    & 3.1&	5.8&	0.03&	0.041&	0.50&	0.54&	9.5\\
    +FP*100
    & 2.8&	10.6&	0.011&	0.031&	0.80&	0.83&	14\\
    \bottomrule
  \end{tabu}
  \label{suppl_tab:delta_res}

  \end{center}

\begin{center}
  \centering
  \captionof{table}{\label{suppl_tab:metrics}Description of error metrics used for evaluation:}
  \begin{tabu} to 0.9\linewidth{ X[2.0c] | X[10.0l] }
    \toprule
    FP/FN & false positive/negative edge\\
    IS & identity switch/cross-over of tracks\\
    FP/FNdiv & false positive/negative division\\
    div & sum of division errors\\
    sum & sum of all errors (incl. divisions)\\
    DET & weighted, normalized score over how many false positive detections have to be deleted and false negatives have to be added to get from the prediction to the ground truth.\\
    TRA & similar to DET but with additional terms for the addition, deletion and modification of links between objects\\
  \end{tabu}
\end{center}

\begin{center}
  \captionof{table}{Information about implementation and computational aspects (for more details see \url{https://github.com/funkelab/linajea}):}
  \begin{longtable}{ p{.30\textwidth}  p{.70\textwidth} }
    \toprule
     &  Description\\
    \midrule
    Tracking Network &    4d U-Net, 3 levels; valid padding; starting with 12 feature maps; quadrupled at each level; separable transposed convolutions for upsampling; if anisotropic, don't downsample anisotropic dimension until voxel size across dimensions is roughly isotropic\\
&     Input size (in voxels): [7, 40, 148, 148] (\textit{mskcc-confocal}), [7, 148, 148, 148] (\textit{nih-ls})\\
&     Pixels outside of cell (determined by some roughly estimated radius) trained with loss factor \(0.01\) (\textit{mskcc-confocal})/ \(0.000001\) (\textit{nih-ls})\\
&     At every iteration a random tile (of size input size) is selected: select a random cell; place input tile around it such that the cell is contained in it, at a random location within tile; depending on the location there might be many neighboring cells included, cells with more neighbors will be sampled less often, as they will also be included if their neighbors are sampled; at least 25\% of iterations have to contain a division\\
&     Adam optimizer; batch size 1; learning rate 0.00005; weighted MSE loss; automatic mixed precision\\
&     Trained for 400k iterations (last checkpoint used for prediction)\\
&     Stochastic weight averaging every 1k after 50k\\
&     Augmentations: elastic, scale, intensity, flipping, shift (of center frame)\\
&     During inference maxima with a score below \(0.2\) are discarded\\
\midrule
    Cell State Network &    3d ResNet18\\
&     Input size (in voxels): [5, 8, 64, 64] (\textit{mskcc-confocal}), [5, 32, 32, 32] (\textit{nih-ls})\\
&     Adam optimizer; batch size 64 (\textit{mskcc-confocal}), 16 (\textit{nih-ls}); learning rate 0.0005, 0.000005 after 20k iterations; cross entropy or focal loss; automatic mixed precision\\
&     Dropout; no batch normalization; global average pooling\\
&     Trained for 60k iterations; best checkpoint selected via validation\\
&     No test time augmentation\\
&     No stochastic weight averaging\\
&     trained on one embryo; validated on a second embryo; inference run on cells predicted by the tracking network on third embryo (same combination for both networks)\\
&     Augmentations: elastic, scale, intensity, flipping, jitter, noise (s\&p, speckle, only for \textit{mskcc-confocal})\\
\midrule
    Postprocessing &    Block-wise processing: tiled only temporally (multiple whole frames) \\
&     After a reasonable range for values for weights had been determined validate on/search a set of 50 different combinations per experiment to find optimal ones\\
&     Ground truth annotation and predicted cell have to be closer than 15 (in world units/isotropic size) to be able to count as a match\\
&     Set of weights selected that results in the lowest sum of errors on the validation set\\
\midrule
Runtime  & U-Net: training around 7 days, prediction a couple of hours\\
(very roughly,  &     ResNet: training 1 day, prediction 1 hour\\
on a single node, &     Generating edges: 5min\\
steps after training &     Solving with one set of ILP weights: 15min\\
can be parallelized) &     Evaluation: 1-5min (depending on what scores should be computed)\\
\midrule
Hardware & Trained on one machine with one V100 (for anisotropic data a smaller GPU is sufficient, too)\\
&     GPU only needed for training and prediction\\
&     Weight search can be parallelized trivially and profits from many CPU cores\\
    \bottomrule
  \end{longtable}
  \label{suppl_tab:impl}

\end{center}

\begin{minipage}{\textwidth}
  \centering
  \hspace*{\fill}%
  \includegraphics[width=0.45\textwidth]{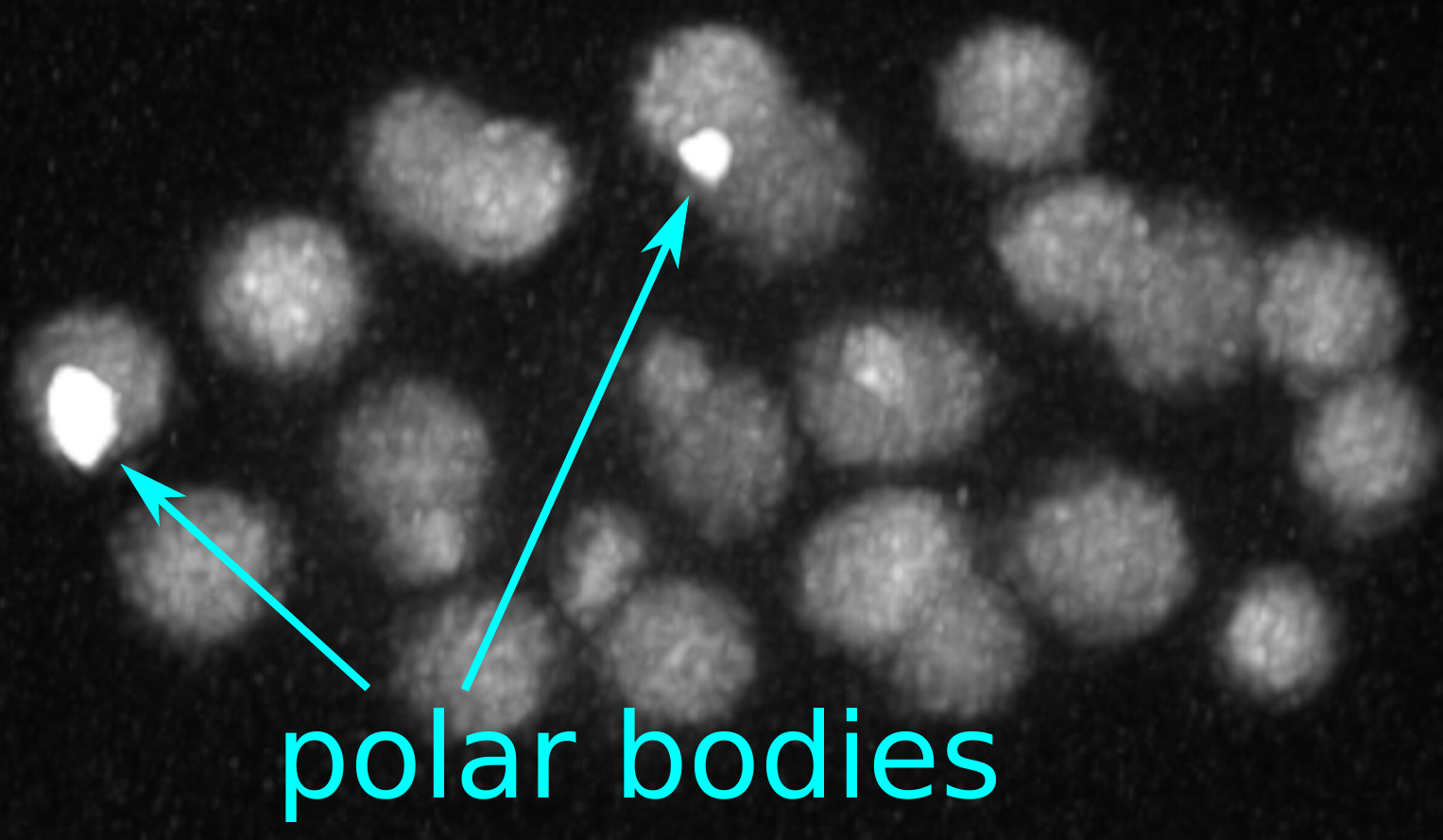}%
  \hfill%
  \includegraphics[width=0.45\textwidth]{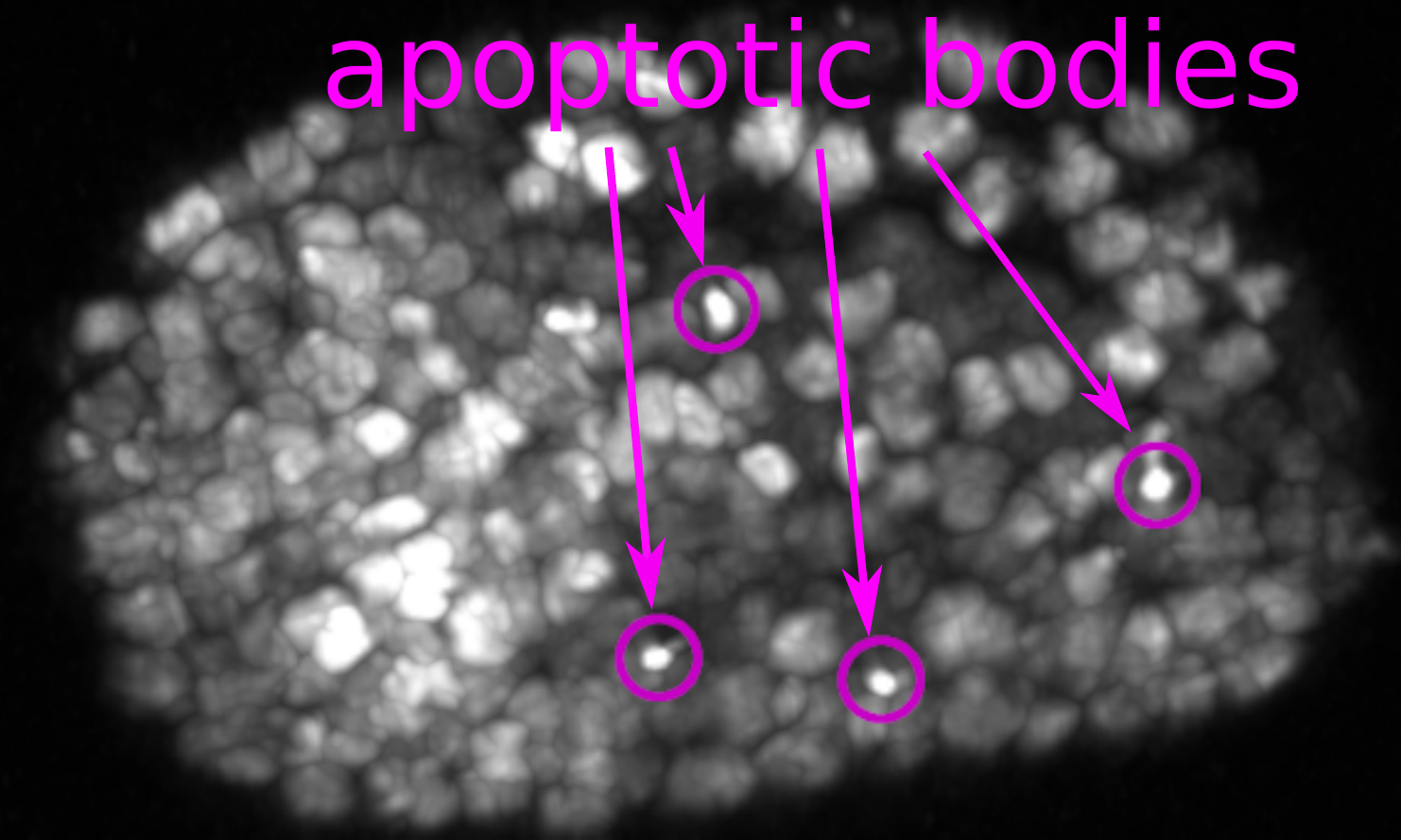}%
  \hspace*{\fill}%
  \captionof{figure}{\label{fig:pol_apo_bodies}Examples of polar and apoptotic bodies in \textit{C. elegans}}
\end{minipage}

\begin{minipage}{\textwidth}
  \centering
  \includegraphics[width=\textwidth]{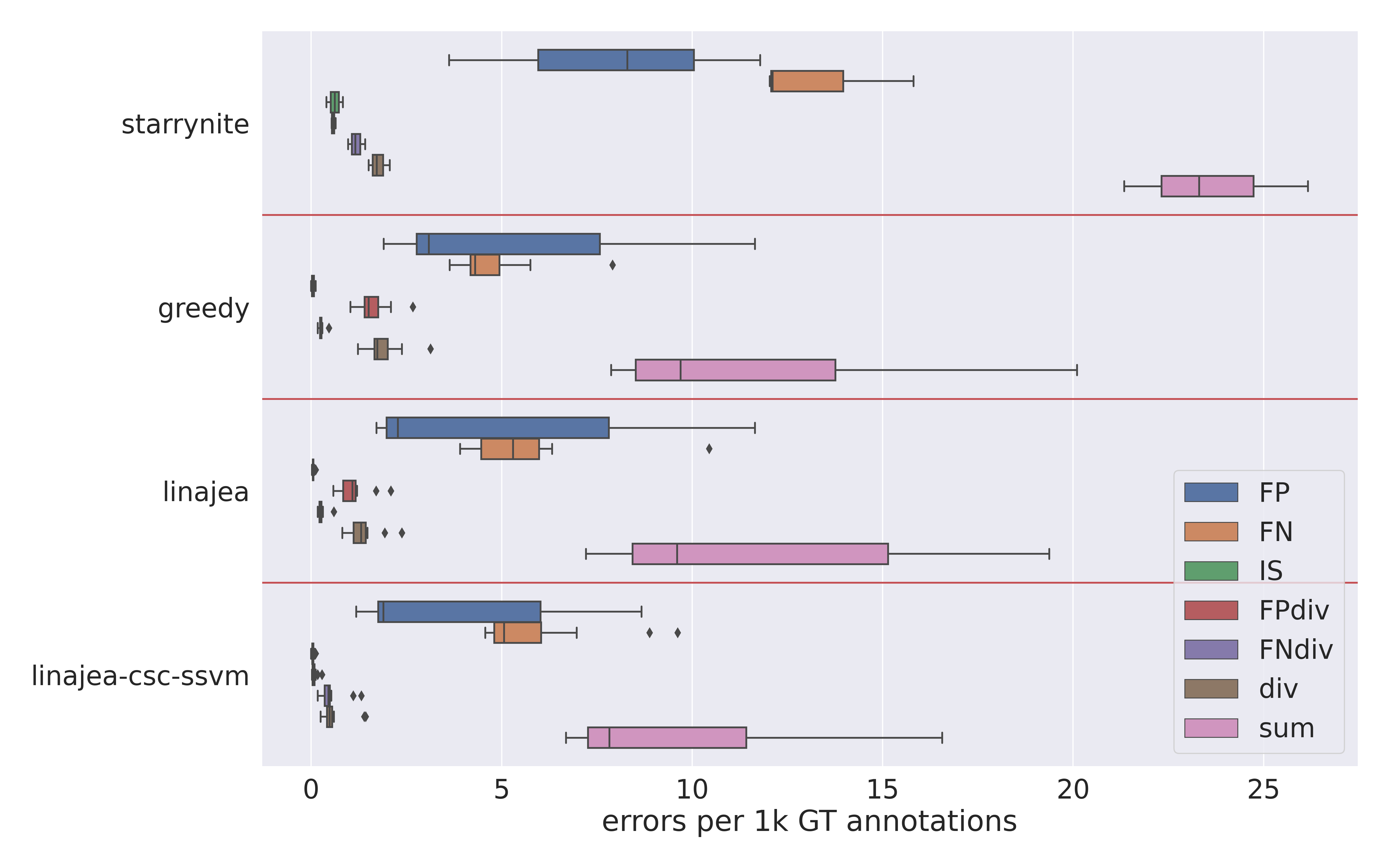}
  \captionof{figure}{\label{fig:error_plot}Box plot of the errors of the different approaches on \textbf{mskcc-confocal}.
    Starrynite~\cite{bao06_autom_cell_lineag_tracin_caenor_elegan} is an often used method in the analysis of \textit{C. elegans}. \textit{greedy} refers to the method without the ILP. \textit{linajea} matches the prior work~\cite{Malin-Mayor2021_autom_recon_of_whole_embry_cell_linea_by_learn_from_spars_annot}. \textit{linajea+csc+ssvm} is our full method with automatically determined ILP weights. Each step lowers the number of errors. \textit{greedy} lowers especially the number of \textit{FP} and \textit{FN} edges, not as much the number of false divisions. The ILP on its own (\textit{linajea}) can already lower the number of false divisions a bit, but the inclusion of the classifier in \textit{linajea+csc+ssvm} lowers them drastically. For the quantitative numbers see Table~\ref{suppl_tab:ablation}.
  }
\end{minipage}

\begin{minipage}{\textwidth}
  \centering
  \hspace*{\fill}%
  \includegraphics[width=0.45\textwidth]{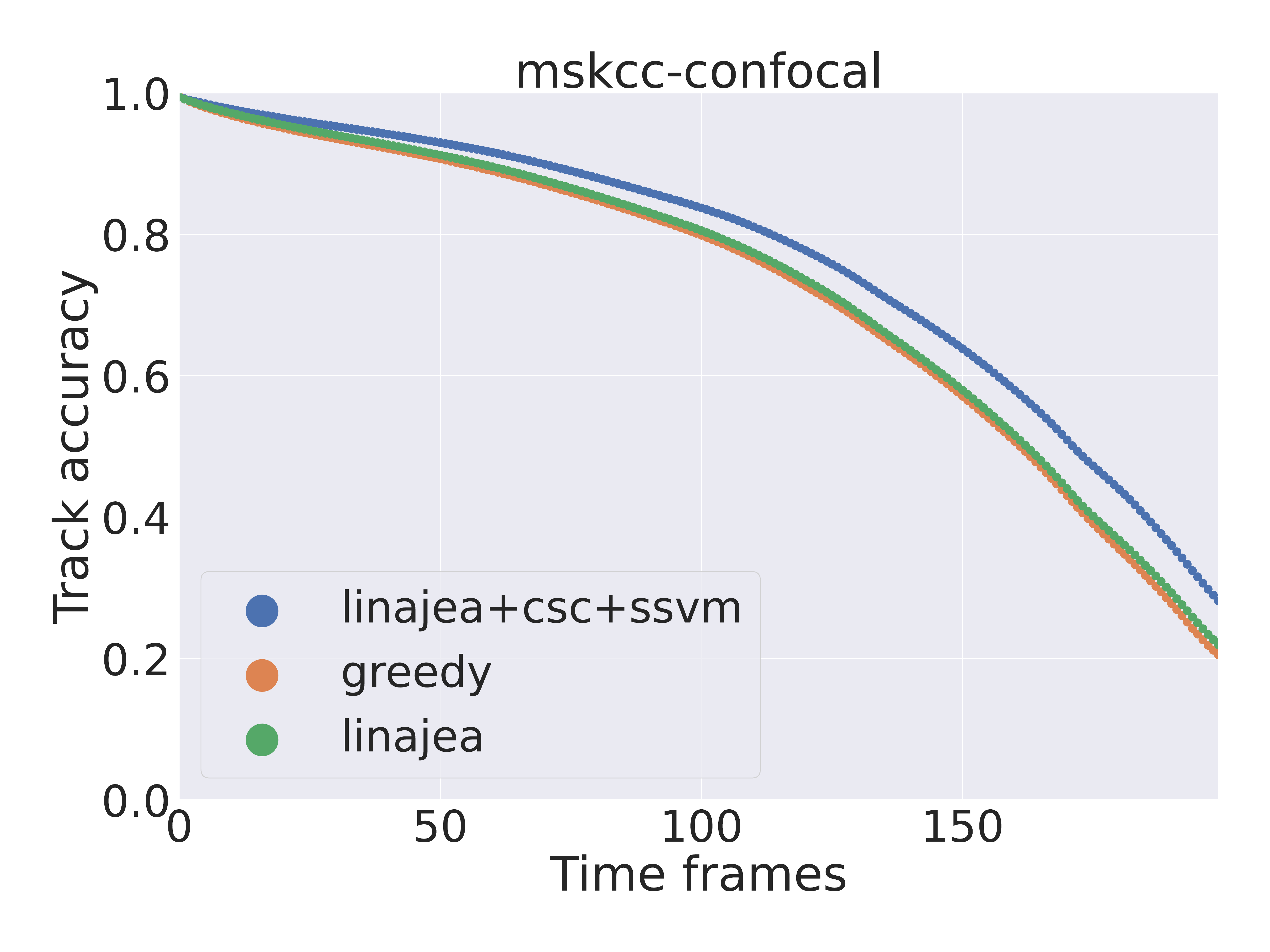}%
  \hfill%
  \includegraphics[width=0.45\textwidth]{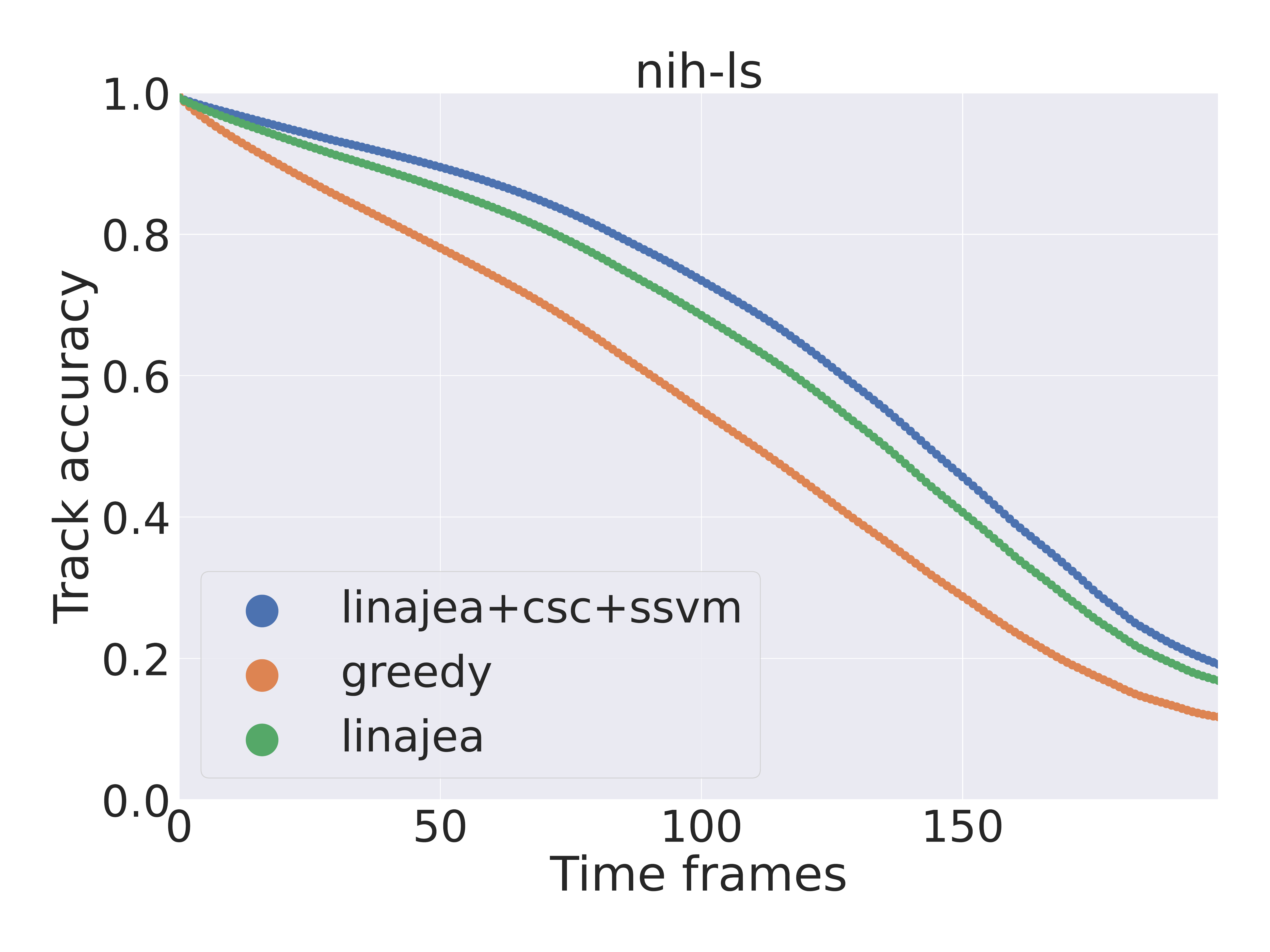}%
  \hspace*{\fill}%
  \captionof{figure}{\label{fig:corr_segments}Fraction of correct tracklets of varying length for \textit{mskcc-confocal} (left, improvement of 3 percentage points at 100 frames and 6 at 200 frames of linajea+csc+ssvm over linajea) and \textit{nih-ls} (right, improvement of 5 percentage points at 100 frames and 2 at 200 frames over linajea), \(p<0.01\). Computed using a sliding window approach; for each window size in the range \([1, 200]\) we determine for every possible tracklet of that length if it is error-free or not and compute the fraction of fully correct ones.}
\end{minipage}

\begin{minipage}{\textwidth}
  \centering
  \includegraphics[width=\textwidth]{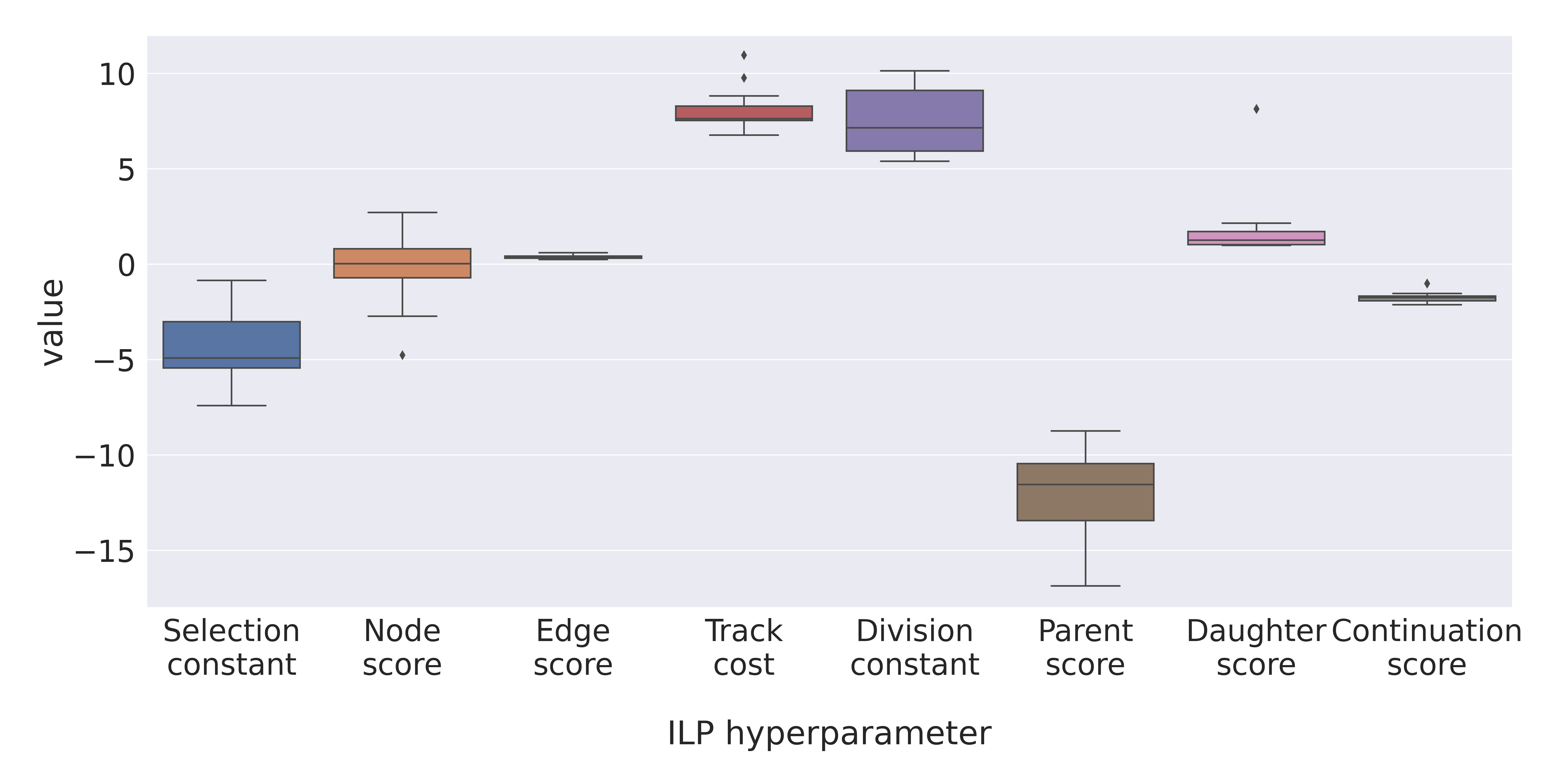}
  \captionof{figure}{\label{fig:ssvm_param_dist}Box and whisker plot of the distribution of the automatically determined ILP weights over the 18 experimental runs of the \textbf{mskcc-confocal} dataset. The sSVM finds similar values for each respective candidate graph and with a similar ratio to each other.}
\end{minipage}


%% file: main.bbl
\begin{thebibliography}{10}
\providecommand{\url}[1]{\texttt{#1}}
\providecommand{\urlprefix}{URL }
\providecommand{\doi}[1]{https://doi.org/#1}

\bibitem{bao06_autom_cell_lineag_tracin_caenor_elegan}
Bao, Z., Murray, J.I., Boyle, T., Ooi, S.L., Sandel, M.J., Waterston, R.H.:
  Automated cell lineage tracing in caenorhabditis elegans. Proceedings of the
  National Academy of Sciences  \textbf{103}(8),  2707--2712 (2006).
  \doi{10.1073/pnas.0511111103}

\bibitem{cao2020establishment}
Cao, J., Guan, G., Ho, V.W.S., Wong, M.K., Chan, L.Y., Tang, C., Zhao, Z., Yan,
  H.: Establishment of a morphological atlas of the caenorhabditis elegans
  embryo using deep-learning-based 4d segmentation. Nature communications
  \textbf{11}(1) (2020)

\bibitem{ccicek16_u_net}
Cicek, O., Abdulkadir, A., Lienkamp, S.S., Brox, T., Ronneberger, O.: 3d u-net:
  Learning dense volumetric segmentation from sparse annotation. CoRR  (2016),
  \url{http://arxiv.org/abs/1606.06650v1}

\bibitem{funke_conv4d}
{Funke, Jan}: {4d convolution implementation} (2018)

\bibitem{guignard20_conta_area_depen_cell_commu_and_the_morpho_invar_of_ascid_embry}
Guignard, L., Fi{\'u}za, U.M., Leggio, B., Laussu, J., Faure, E., Michelin, G.,
  Biasuz, K., Hufnagel, L., Malandain, G., Godin, C., Lemaire, P.: Contact
  area{\textendash}dependent cell communication and the morphological
  invariance of ascidian embryogenesis. Science  \textbf{369}(6500) (2020).
  \doi{10.1126/science.aar5663}

\bibitem{he15_deep_resid_learn_image_recog}
He, K., Zhang, X., Ren, S., Sun, J.: Deep residual learning for image
  recognition. CoRR  (2015), \url{http://arxiv.org/abs/1512.03385v1}

\bibitem{hofener2018deep}
H{\"o}fener, H., Homeyer, A., Weiss, N., Molin, J., Lundstr{\"o}m, C.F., Hahn,
  H.K.: Deep learning nuclei detection: A simple approach can deliver
  state-of-the-art results. Computerized Medical Imaging and Graphics
  \textbf{70},  43--52 (2018)

\bibitem{joachims09_predi_struc_objec_with_suppo_vecto_machin}
Joachims, T., Hofmann, T., Yue, Y., Yu, C.N.: Predicting structured objects
  with support vector machines. Commun. ACM  \textbf{52}(11),  97–104 (nov
  2009). \doi{10.1145/1592761.1592783},
  \url{https://doi.org/10.1145/1592761.1592783}

\bibitem{jug2014optimal}
Jug, F., Pietzsch, T., Kainm{\"u}ller, D., Funke, J., Kaiser, M., van Nimwegen,
  E., Rother, C., Myers, G.: Optimal joint segmentation and tracking of
  escherichia coli in the mother machine. In: Bayesian and graphical Models for
  Biomedical Imaging. Springer (2014)

\bibitem{keller2010fast}
Keller, P.J., Schmidt, A.D., Santella, A., Khairy, K., Bao, Z., Wittbrodt, J.,
  Stelzer, E.H.: Fast, high-contrast imaging of animal development with scanned
  light sheet--based structured-illumination microscopy. Nature methods
  \textbf{7}(8),  637--642 (2010)

\bibitem{krzic2012multiview}
Krzic, U., Gunther, S., Saunders, T.E., Streichan, S.J., Hufnagel, L.:
  Multiview light-sheet microscope for rapid in toto imaging. Nature methods
  \textbf{9}(7) (2012)

\bibitem{li2019_syste_probe_and_spati_regul_of_cell_posit_varia_durin_embry}
Li, X., Zhao, Z., Xu, W., Fan, R., Xiao, L., Ma, X., Du, Z.: Systems properties
  and spatiotemporal regulation of cell position variability during
  embryogenesis. Cell Reports  \textbf{26}(2),  313--321.e7 (2019).
  \doi{https://doi.org/10.1016/j.celrep.2018.12.052}

\bibitem{magnusson15_global_linkin_cell_track_using_viter_algor}
Magnusson, K.E.G., Jalden, J., Gilbert, P.M., Blau, H.M.: Global linking of
  cell tracks using the viterbi algorithm. IEEE Transactions on Medical Imaging
   \textbf{34}(4),  911--929 (2015). \doi{10.1109/tmi.2014.2370951}

\bibitem{Malin-Mayor2021_autom_recon_of_whole_embry_cell_linea_by_learn_from_spars_annot}
Malin-Mayor, C., Hirsch, P., Guignard, L., McDole, K., Wan, Y., Lemon, W.C.,
  Keller, P.J., Preibisch, S., Funke, J.: Automated reconstruction of
  whole-embryo cell lineages by learning from sparse annotations. bioRxiv
  (2021). \doi{10.1101/2021.07.28.454016}

\bibitem{matula15_cell_track_accur_measur_based}
Matula, P., Ma{\v{s}}ka, M., Sorokin, D.V., Matula, P., de~Sol{\'o}rzano, C.O.,
  Kozubek, M.: Cell tracking accuracy measurement based on comparison of
  acyclic oriented graphs. PLOS ONE  \textbf{10}(12),  e0144959 (2015).
  \doi{10.1371/journal.pone.0144959}

\bibitem{medeiros2021_multisc_light_sheet_organ_imagi_frame}
de~Medeiros, G., Ortiz, R., Strnad, P., Boni, A., Maurer, F., Liberali, P.:
  Multiscale light-sheet organoid imaging framework. bioRxiv  (2021).
  \doi{10.1101/2021.05.12.443427}

\bibitem{Moyle2021}
Moyle, M.W., Barnes, K.M., Kuchroo, M., Gonopolskiy, A., Duncan, L.H.,
  Sengupta, T., Shao, L., Guo, M., Santella, A., Christensen, R., Kumar, A.,
  Wu, Y., Moon, K.R., Wolf, G., Krishnaswamy, S., Bao, Z., Shroff, H., Mohler,
  W.A., Col{\'o}n-Ramos, D.A.: Structural and developmental principles of
  neuropil assembly in c. elegans. Nature  \textbf{591}(7848) (2021).
  \doi{10.1038/s41586-020-03169-5}

\bibitem{murray08_autom_analy_embry_gene_expres}
Murray, J.I., Bao, Z., Boyle, T.J., Boeck, M.E., Mericle, B.L., Nicholas, T.J.,
  Zhao, Z., Sandel, M.J., Waterston, R.H.: Automated analysis of embryonic gene
  expression with cellular resolution in c. elegans. Nature Methods
  \textbf{5}(8) (2008). \doi{10.1038/nmeth.1228}

\bibitem{pogan2020_diffe_of_black_combi_solve}
Pogančić, M.V., Paulus, A., Musil, V., Martius, G., Rolinek, M.:
  Differentiation of blackbox combinatorial solvers. In: International
  Conference on Learning Representations (2020),
  \url{https://openreview.net/forum?id=BkevoJSYPB}

\bibitem{ronneberger15_u_net}
Ronneberger, O., Fischer, P., Brox, T.: U-net: Convolutional networks for
  biomedical image segmentation. CoRR  (2015),
  \url{http://arxiv.org/abs/1505.04597v1}

\bibitem{santella14_semi_local_neigh_based_framew}
Santella, A., Du, Z., Bao, Z.: A semi-local neighborhood-based framework for
  probabilistic cell lineage tracing. BMC Bioinformatics  \textbf{15}(1)
  (2014). \doi{10.1186/1471-2105-15-217}

\bibitem{schiegg2013conservation}
Schiegg, M., Hanslovsky, P., Kausler, B.X., Hufnagel, L., Hamprecht, F.A.:
  Conservation tracking. In: Proceedings of the IEEE International Conference
  on Computer Vision (2013)

\bibitem{sugawara2021_track_cell_linea_in_3d_by_incre_deep_learn}
Sugawara, K., Cevrim, C., Averof, M.: Tracking cell lineages in 3d by
  incremental deep learning. bioRxiv  (2021). \doi{10.1101/2021.02.26.432552}

\bibitem{teo10_bundl_metho_for_regul_risk_minim}
Teo, C.H., Vishwanthan, S., Smola, A.J., Le, Q.V.: Bundle methods for
  regularized risk minimization. Journal of Machine Learning Research
  \textbf{11}(10),  311--365 (2010),
  \url{http://jmlr.org/papers/v11/teo10a.html}

\bibitem{ulman17_objec_compar_cell_track_algor}
Ulman, V., Ma{\v{s}}ka, M., Magnusson, K.E.G., Ronneberger, O., Haubold, C.,
  Harder, N., Matula, P., Matula, P., Svoboda, D., Radojevic, M., Smal, I.,
  Rohr, K., Jald{\'e}n, J., Blau, H.M., Dzyubachyk, O., Lelieveldt, B., Xiao,
  P., Li, Y., Cho, S.Y., Dufour, A.C., Olivo-Marin, J.C., Reyes-Aldasoro, C.C.,
  Solis-Lemus, J.A., Bensch, R., Brox, T., Stegmaier, J., Mikut, R., Wolf, S.,
  Hamprecht, F.A., Esteves, T., Quelhas, P., Demirel, {\"O}., Malmstr{\"o}m,
  L., Jug, F., Tomancak, P., Meijering, E., Mu{\~n}oz-Barrutia, A., Kozubek,
  M., de~Solorzano, C.O.: An objective comparison of cell-tracking algorithms.
  Nature Methods  \textbf{14}(12),  1141--1152 (2017). \doi{10.1038/nmeth.4473}

\bibitem{Weigert2018_CARE}
Weigert, M., Schmidt, U., Boothe, T., M{\"u}ller, A., Dibrov, A., Jain, A.,
  Wilhelm, B., Schmidt, D., Broaddus, C., Culley, S., Rocha-Martins, M.,
  Segovia-Miranda, F., Norden, C., Henriques, R., Zerial, M., Solimena, M.,
  Rink, J., Tomancak, P., Royer, L., Jug, F., Myers, E.W.: Content-aware image
  restoration: pushing the limits of fluorescence microscopy. Nature Methods
  \textbf{15}(12),  1090--1097 (Dec 2018). \doi{10.1038/s41592-018-0216-7}

\bibitem{Wolff2018}
Wolff, C., Tinevez, J.Y., Pietzsch, T., Stamataki, E., Harich, B., Guignard,
  L., Preibisch, S., Shorte, S., Keller, P.J., Tomancak, P., Pavlopoulos, A.:
  Multi-view light-sheet imaging and tracking with the mamut software reveals
  the cell lineage of a direct developing arthropod limb. eLife  \textbf{7},
  e34410 (mar 2018). \doi{10.7554/eLife.34410}

\end{thebibliography}
